\documentclass{article}
\usepackage{ijcai25}

\usepackage{times}
\usepackage{soul}
\usepackage{url}
\usepackage[hidelinks]{hyperref}
\usepackage[utf8]{inputenc}
\usepackage[small]{caption}
\usepackage{graphicx}
\usepackage{amsmath}
\usepackage{amsthm}
\usepackage{booktabs}
\usepackage{algorithm}
\usepackage{algorithmic}
\usepackage[switch]{lineno}
\usepackage{float}
\usepackage{adjustbox}
\usepackage{array}
\usepackage{multirow}
\usepackage{amssymb}
\usepackage{bm} % For bold math symbols
\usepackage{enumitem}
\usepackage{natbib}
\usepackage{bibentry}

\title{FedAlign: Federated Domain Generalization with Cross-Client Feature Alignment}

\author{
    Sunny Gupta\textsuperscript{\rm 1}, 
    Vinay Sutar\textsuperscript{\rm 2}, 
    Varunav Singh\textsuperscript{\rm 3}, 
    Amit Sethi\textsuperscript{\rm 4} \\
    \textsuperscript{\rm 1,2,3}Indian Institute of Technology Bombay, India \\
    \{sunnygupta, 21d070078, 21d070086, asethi\}@iitb.ac.in
}

\begin{document}

\maketitle

\begin{abstract}
\textit{Federated Learning (FL)} offers a decentralized paradigm for collaborative model training without direct data sharing, yet it poses unique challenges for \textit{Domain Generalization (DG)}, including strict privacy constraints, non-i.i.d. local data, and limited domain diversity. We introduce \textbf{FedAlign}, a lightweight, privacy-preserving framework designed to enhance DG in federated settings by simultaneously increasing feature diversity and promoting domain invariance. First, a cross-client feature extension module broadens local domain representations through domain-invariant feature perturbation and selective cross-client feature transfer, allowing each client to safely access a richer domain space. Second, a dual-stage alignment module refines global feature learning by aligning both feature embeddings and predictions across clients, thereby distilling robust, domain-invariant features. By integrating these modules, our method achieves superior generalization to unseen domains while maintaining data privacy and operating with minimal computational and communication overhead.
\end{abstract}

\section{Introduction}
Conventional machine learning techniques are built on the assumption that training and test data are identically and independently distributed (IID). However, this assumption is often violated in real-world applications where models frequently encounter Out-of-Distribution (OOD) data, leading to significant performance degradation on unseen domains \citep{recht2019imagenet}. For instance, a model trained on cartoon images may fail to generalize to sketches due to domain shifts. \textbf{Domain Generalization (DG)} aims to address this limitation by equipping models with the ability to generalize effectively to unseen data distributions \citep{zhou2022domain}.

\begin{figure}[t] % [t] places the figure at the top of the column
    \centering
    \includegraphics[width=\linewidth]{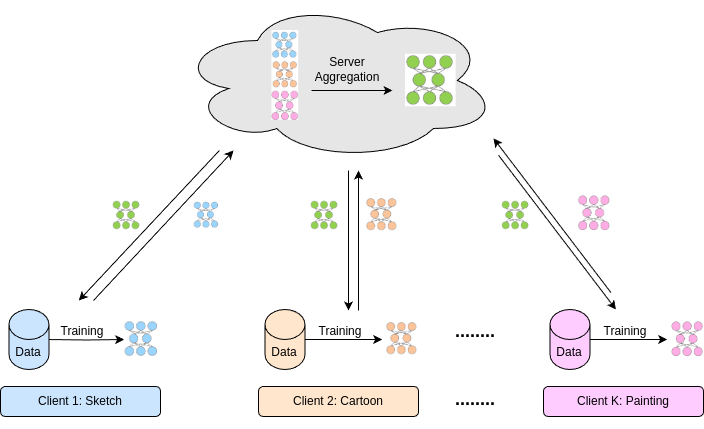} % Adjust width to fit the column
    \caption{Illustration of the typical scenario in FL. Each client
    contains data from a unique domain, and the test domain (Photo)
    differs from all domains present on the clients.}
    \label{fig:fedalign}
\end{figure}

Despite the promise of DG, many existing approaches depend on centralized datasets, a condition that is infeasible in scenarios where data is distributed across multiple clients. \textbf{Federated Learning (FL)} \citep{mcmahan2017communication} provides a decentralized alternative by enabling collaborative model training without exposing raw data. However, integrating DG within FL poses unique challenges, including limited domain diversity at the client level and stringent privacy constraints inherent in decentralized environments.

\begin{figure*}[t]
    \centering
    \includegraphics[width=\textwidth, keepaspectratio]{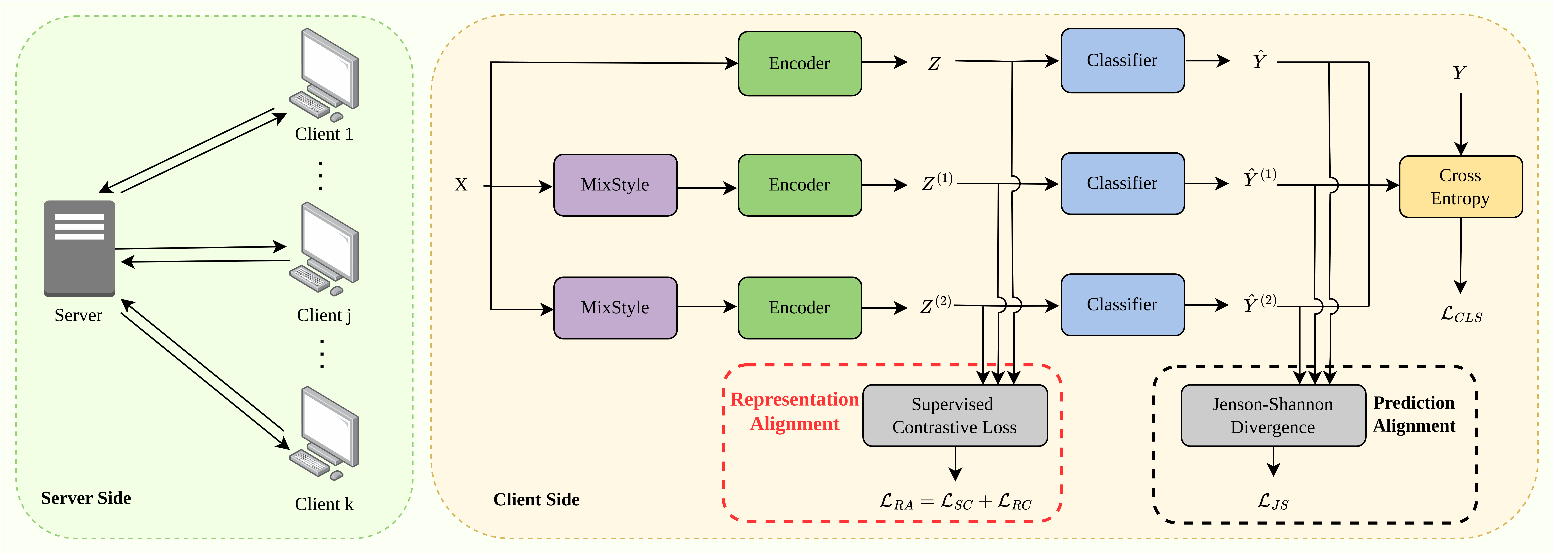}
    \caption{Overview of FedAlign: Clients share local model parameters and sample statistics with the server, which aggregates and redistributes them. Local training incorporates feature augmentation, representation alignment, and prediction alignment to enhance domain-invariant feature learning.}
    \label{fig:FedAlign}
\end{figure*}

\textit{Federated Domain Generalization (FDG)} focuses on learning domain-invariant features—label-relevant attributes that remain stable across diverse domains. Current FDG approaches predominantly employ two main strategies:

These methods use adversarial objectives to align representations \citep{micaelli2019zero, peng2019federated, xu2023federated, zhang2021federated}, but they often incur high computational overhead and can suffer from training instabilities such as model collapse \citep{arjovsky2017wasserstein}.

By aligning features across clients, this line of work aims to mitigate domain discrepancies \citep{nguyen2022fedsr, yao2022federated, zhang2021federated}. However, limited domain diversity at the client level and strict privacy constraints can hinder alignment effectiveness at scale.

An alternative solution space involves \textit{Federated Style Transfer} \citep{yang2020fda, yoon2021fedmix}, which augments local data diversity via techniques like AdaIN \citep{huang2017arbitrary} and CycleGAN \citep{zhu2017unpaired}. While effective at generating domain-varied samples, these approaches often demand additional models for feature extraction and high-dimensional embedding exchanges, resulting in: Substantial communication overhead, Heightened privacy risks \citep{chen2023federated, park2024stablefdg}, Limited improvements in domain-invariant feature learning.

To address these challenges, we propose \textbf{FedAlign}, a federated domain generalization framework:

FedAlign introduces a novel feature-sharing mechanism that enriches each client’s domain exposure without revealing raw data. This strategy perturbs domain-invariant features and redistributes them across clients in a privacy-preserving manner, broadening the effective training distribution while upholding confidentiality.

A two-step alignment process ensures consistent performance across varied domains: Supervised Contrastive Loss encourages representations of samples with identical labels to converge, effectively reducing intra-class variance across domains. Jensen–Shannon Divergence enforces prediction consistency by aligning output distributions for both original and perturbed data, further bolstering out-of-distribution robustness.

Unlike adversarial training or style transfer-based methods, FedAlign’s lightweight feature-sharing mechanism imposes negligible additional overhead, making it well-suited for large-scale FL systems.

By focusing on privacy-preserving feature transfers and a dual-stage alignment of representations and predictions, FedAlign addresses the critical limitations of existing FDG methods specifically, the interplay of limited local data, insufficient domain diversity, and strict privacy constraints.

\section{Related Work}
\subsection{Representation Alignment}
Another prominent line of research in \textit{Domain Generalization (DG)} focuses on representation alignment—reducing domain-specific variations by aligning feature distributions across multiple domains. Notable examples include:

Approaches like DANN \citep{ganin2015unsupervised, ganin2016domain, gong2019dlow} deploy a domain classifier to guide alignment, ensuring the extracted features are domain-invariant. By training the feature extractor and domain classifier in an adversarial manner, these methods successfully mitigate domain discrepancies.

CORAL \citep{sun2016deep} aligns second-order statistics (e.g., covariance matrices) between source and target feature distributions, thereby reducing mismatches in feature representations.

Methods grounded in Maximum Mean Discrepancy (MMD) \citep{tzeng2014deep, wang2018visual, wang2020transfer} leverage kernel-based metrics to align representations across domains, promoting more universal feature embeddings.

Although these alignment techniques have demonstrated improved performance on unseen domains, they commonly assume centralized access to all training domains. Such an assumption conflicts with the privacy-preserving requirements of \textit{Federated Learning (FL)}, where data cannot be directly exchanged among clients or with a central server. Consequently, adapting representation alignment methods to FL necessitates innovative strategies that ensure robust domain generalization without violating data privacy constraints.

\subsection{Style Transfer}
A range of style transfer-based domain generalization (DG) methods \citep{volpi2019addressing, volpi2018generalizing, xu2020robust} aim to enrich domain diversity, thereby improving model robustness on unseen target domains. These approaches can be broadly separated into two main categories:

In the first category, generative models are employed to synthesize data with diverse styles \citep{palakkadavath2024domain, robey2021model}. By enhancing variability in color, texture, and other visual attributes, these methods reduce reliance on domain-specific features. However, generative modeling often demands substantial computational resources and can encounter training instability—including model collapse in adversarial training—thereby jeopardizing convergence and overall performance.

% \begin{flushleft}
% \begin{figure}[htbp]
\begin{algorithm}[H]
\caption{FedAlign}
\label{alg:fedalign}

\textbf{Input:} Client datasets \( \{D_k \mid k = 1, \dots, K\} \), where \( D_k = \{(x_i, y_i)\}_{i=1}^{n_k} \). \\
Global model \( f = g \circ h \), where \( h(\cdot) \) is the encoder and \( g(\cdot) \) is the classifier. \\
Number of communication rounds \( T \), local epochs \( E \), and learning rate \( \eta \).

\begin{algorithmic}[1]
\STATE Initialize global model parameters \( \theta_0 \).
\STATE \textbf{Server Side:}
\FOR{\( t = 1, \dots, T \)}
    \STATE Select a subset of clients \( C_t \) to participate.
    \STATE Broadcast global parameters \( \theta_t \) to selected clients.
    \FOR{\( k \in C_t \)}
        \STATE Receive updated client parameters \( \theta_{k, t+1} \).
    \ENDFOR
    \STATE Update global parameters:
    \[
    \theta_{t+1} = \frac{1}{N} \sum_{k \in C_t} n_k \theta_{k, t+1}, \quad N = \sum_{k \in C_t} n_k.
    \]
\ENDFOR

\STATE \textbf{Client Side:}
\STATE Input global parameters \( \theta_t \).
\FOR{\( e = 1, \dots, E \)}
    \FOR{a batch \( X \in \mathbb{R}^{B \times C \times H \times W} \)}
        \STATE Generate augmented batches:
        \[
        X^{(1)} = M(X), \quad X^{(2)} = M(X).
        \]
        \STATE Compute representations and predictions:
        \[
        Z, Z^{(1)}, Z^{(2)} = h(X), h(X^{(1)}), h(X^{(2)}),
        \]
        \[
        \hat{Y}, \hat{Y}^{(1)}, \hat{Y}^{(2)} = g(Z), g(Z^{(1)}), g(Z^{(2)}).
        \]
        \STATE Compute losses:
        \[
        L_{\text{CLS}} = \frac{1}{B} \sum_{i=1}^B \ell(\hat{y}_i, y_i),
        \]
        \[
        L_{\text{SC}} = \frac{1}{2} \left( L_{\text{SC}}(Z^{(1)}, Z) + L_{\text{SC}}(Z^{(2)}, Z) \right),
        \]
        \[
        L_{\text{RC}} = \frac{1}{|mix\_feat|} \sum \| h(X) - h(X_{\text{aug}}) \|^2,
        \]
        \[
        L_{\text{RA}} = L_{\text{SC}} + L_{\text{RC}},
        \]
        \[
        L_{\text{JS}} = \frac{1}{3} \left( \text{KL}(\hat{Y} \| \overline{Y}) + \text{KL}(\hat{Y}^{(1)} \| \overline{Y}) + \text{KL}(\hat{Y}^{(2)} \| \overline{Y}) \right),
        \]
        where \( \overline{Y} = \frac{1}{3} (\hat{Y} + \hat{Y}^{(1)} + \hat{Y}^{(2)}) \).
        \STATE Compute total loss:
        \[
        L = L_{\text{CLS}} + \lambda_1 L_{\text{RA}} + \lambda_2 L_{\text{JS}}.
        \]
        \STATE Update local model with gradient step on \( L \) using \( \eta \).
    \ENDFOR
\ENDFOR
\STATE Return \( \theta_{k, t+1} \) to the server.

\textbf{Output:} Global model \( \theta_T \) after \( T \) rounds of communication.
\end{algorithmic}
\end{algorithm}
% \end{figure}
% \end{flushleft}

A second line of work leverages data augmentation techniques such as MixStyle \citep{zhou2021domain} and Mixup \citep{zhang2018mixup}. Instead of synthesizing entirely new samples, these methods manipulate existing data to boost intra-batch diversity. Specifically:
MixStyle interpolates channel-wise style statistics within a batch, promoting domain-invariant feature learning. Mixup merges data samples and their corresponding labels to expand the decision boundary.

Compared to generative approaches, augmentation-based methods are more computationally efficient and inherently free from adversarial training instability, rendering them particularly suitable for large-scale DG applications.

\subsection{Federated Domain Generalization}

Most existing \textit{Federated Domain Generalization (FDG)} methods aim to learn domain-invariant representations across heterogeneous clients. Common strategies include federated adversarial learning and federated representation alignment, yet each approach faces notable challenges:

Approaches like FedADG \citep{zhang2021federated} employ a global discriminator to extract universal feature representations while preserving local data privacy. Although this technique can mitigate domain discrepancies, it often incurs high computational costs and risks training instability, including potential model collapse.

Methods such as FedSR \citep{nguyen2022fedsr} harness L2-norm and conditional mutual information regularization to align feature distributions among clients. These strategies, however, struggle in large-scale federated learning settings, particularly due to limited domain diversity at the individual client level. As a result, models may fail to robustly capture the full variability needed for strong out-of-distribution generalization.

To alleviate the challenge of limited local data diversity, CCST \citep{chen2023federated} incorporates cross-client style transfer based on AdaIN \citep{huang2017arbitrary}. By generating synthetic samples styled after other domains, CCST expands the effective training distribution. However, this method relies heavily on pre-trained VGG networks \citep{simonyan2014very} for feature extraction and image reconstruction, demanding the transmission of high-dimensional representations. This not only introduces significant communication and computational overhead but also raises privacy risks, as intercepted data could be used to reconstruct original samples \citep{li2021survey, mothukuri2021survey}. Moreover, relying on a pre-trained network can partially contradict domain generalization principles if the target domain is inadvertently included in the pre-training dataset.

Some FDG methods adopt alternative optimization or aggregation mechanisms to promote generalization across domains:
\begin{itemize}
    \item \textbf{FedIIR} \citep{guo2023out} aligns client gradients to implicitly learn domain-invariant relationships, improving out-of-distribution generalization.
    \item \textbf{GA} \citep{zhang2023federated} adjusts aggregation weights dynamically to minimize performance disparities among clients, boosting generalization.
\end{itemize}

Despite these contributions, they do not demonstrate consistent superiority over other FDG approaches in empirical evaluations, highlighting the persistent performance and scalability challenges in FDG research.

Overall, while these methods have achieved notable progress, they often neglect the intertwined constraints of limited data volume and restricted domain diversity at the client level. Their reliance on high-dimensional data exchange or computationally expensive adversarial training further underscores the need for more efficient, privacy-preserving, and robust FDG solutions—an issue that FedAlign seeks to address.

\section{Methodology}
\subsection{Preliminary}
\textbf{Federated Domain Generalization (FDG)} aims to train models collaboratively across multiple clients, where each client holds data from distinct domains. The goal is to develop a global model that generalizes effectively to unseen target domains without direct access to their data. Let \( X \) and \( Y \) denote the input and target spaces, respectively. Consider \( M \) source domains:

\begin{equation}
S_{\text{source}} = \{ S_i \mid i = 1, 2, \dots, M \},
\end{equation}

with each domain sampled from a unique joint distribution \( P_i(x, y) \), where \( x \in X \) and \( y \in Y \). These distributions differ significantly across domains, such that \( P_i(x, y) \neq P_j(x, y) \) for \( i \neq j \), reflecting real-world domain shifts in data distribution.

In a federated learning setting, data from each domain \( S_i \) is distributed across \( K \) clients, denoted as \( D_k \subset S_i \). Each client performs local training on its private dataset and communicates only model updates or minimal statistics with a central server to preserve privacy. The objective of FDG is to collaboratively train a global model \( f: X \to Y \) that minimizes the prediction error on an unseen target domain \( S_{\text{target}} \):

\begin{equation}
\min_f \mathbb{E}_{(x, y) \sim S_{\text{target}}} \left[ \ell(f(x), y) \right],
\end{equation}

where \( \ell(\cdot) \) is a task-specific loss function, such as cross-entropy. Importantly, the unseen target domain \( S_{\text{target}} \) is inaccessible during training, and its joint distribution \( P_{\text{target}}(x, y) \) differs from all source domain distributions \( P_i(x, y) \), i.e., \( P_{\text{target}}(x, y) \neq P_i(x, y) \) for all \( i \in \{1, 2, \dots, M\} \).

% \subsubsection{Federated Parameter Aggregation}

% In FDG, each client locally trains its model on its respective data and shares only the updated model parameters with the central server, preserving the privacy of raw data. After receiving these updates, the server performs parameter aggregation to obtain the updated global model parameters:

% \begin{equation}
% \theta_{t+1} = \frac{1}{N} \sum_{i=1}^{K \cdot M} n_i \theta_i^t,
% \end{equation}

% where:
% \begin{itemize}
%     \item \( \theta_i^t \): Model parameters of client \( i \) at communication round \( t \),
%     \item \( n_i \): Number of samples held by client \( i \),
%     \item \( N = \sum_{i=1}^{K \cdot M} n_i \): Total number of samples across all participating clients.
% \end{itemize}

% The aggregated global parameters \( \theta_{t+1} \) are then redistributed to all clients for the next round of local training. This iterative process continues over multiple communication rounds, allowing the global model to learn collaboratively from domain-specific client data.

\begin{table*}[t] % Ensures the table appears at the top of the page
    \centering
    \adjustbox{max width=\textwidth}{
    \begin{tabular}{lccccc!{\vrule width 0.75pt}ccccc!{\vrule width 0.75pt}ccccc!{\vrule width 0.75pt}ccccc}
        \toprule
        \multirow{2}{*}{Algorithm} 
        & \multicolumn{5}{c!{\vrule width 0.75pt}}{PACS} 
        & \multicolumn{5}{c!{\vrule width 0.75pt}}{OfficeHome} 
        & \multicolumn{5}{c!{\vrule width 0.75pt}}{Caltech-10} 
        & \multicolumn{5}{c}{miniDomainNet} \\
        \cmidrule(lr){2-6} \cmidrule(lr){7-11} \cmidrule(lr){12-16} \cmidrule(lr){17-21}
        & P & A & C & S & Avg. 
        & A & C & P & R & Avg. 
        & A & C & D & W & Avg. 
        & C & P & R & S & Avg. \\
        \midrule
        FedAvg   & 90.30 & 69.95 & 75.70 & 71.80 & 76.10 & 61.50 & 49.00 & 76.50 & 77.04 & 66.10 & \textbf{95.09} & 88.07 & 98.13 & \textbf{93.56} & \textbf{94.02} & 64.13 & 56.89 & 68.23 & 50.47 & 59.19 \\
        FedProx  & 91.21 & 69.84 & 73.15 & 70.87 & 75.92 & 62.43 & 51.09 & 75.47 & 76.89 & 66.21 & 94.57 & 88.33 & 97.45 & 85.76 & 91.53 & 62.57 & 55.96 & 67.11 & 49.71 & 59.62 \\
        FedADG   & 89.52 & 63.54 & 71.45 & 64.32 & 72.20 & 60.58 & 47.15 & 72.19 & 75.84 & 63.73 & 95.09 & 88.07 & 99.36 & 93.56 & 94.02 & 60.17 & 56.34 & 68.45 & 49.82 & 59.04 \\
        GA       & 92.98 & 66.01 & 76.43 & 69.23 & 75.81 & 62.05 & 49.27 & 76.10 & 76.91 & 66.37 & 94.99 & 87.44 & 97.45 & 91.53 & 92.85 & 62.89 & 56.12 & 65.05 & 48.41 & 57.92 \\
        FedSR    & 91.42 & 72.68 & 76.01 & 70.73 & 76.84 & \textbf{63.95} & 50.12 & 76.58 & 77.84 & 66.73 & 93.63 & 87.8 & 96.82 & 87.12 & 91.34 & 67.59 & 61.09 & 68.78 & 57.11 & 63.41 \\
        FedIIR   & 91.53 & 71.25 & 78.61 & 71.78 & 77.89 & 61.64 & 50.14 & 75.12 & 76.83 & 65.89 & 94.57 & 88.16 & 98.73 & 90.17 & 92.38 & 63.01 & 57.23 & 64.79 & 47.58 & 57.96 \\
        CCST     & 89.93 & 76.18 & 75.97 & 78.34 & 79.85 & 60.94 & 50.13 & 76.74 & 77.65 & 66.84 & 93.32 & 83.7 & 92.99 & 88.81 & 89.71 & 60.18 & 57.34 & 67.73 & 49.72 & 58.94 \\
        FedAlign  & \textbf{93.11} & \textbf{80.57} & \textbf{77.94} & \textbf{80.20} & \textbf{82.96} & 61.68 & \textbf{56.17} & \textbf{77.04} & \textbf{77.37} & \textbf{68.03} & 94.78 & \textbf{89.85} & \textbf{98.73} & 91.53 & 93.72 & \textbf{67.83} & \textbf{61.18} & \textbf{69.23} & \textbf{56.80} & \textbf{63.76} \\
        \bottomrule
    \end{tabular}
}

    \caption{Test accuracy on each dataset. These experiments were conducted with an upload ratio of $r = 0.1$. Each algorithm was evaluated three times, and the final results represent the average test accuracy.}
    \label{tab:accuracy}
\end{table*}

\subsection{Framework Overview}

An illustration of the proposed FedAlign framework is provided in Fig.~\ref{fig:FedAlign}, and the detailed algorithmic steps can be found in Algorithm~\ref{alg:fedalign}. Our approach integrates MixStyle-based cross-client feature augmentation with multi-level alignment objectives, enabling more robust domain-invariant feature extraction and improved generalization in federated settings.

\subsubsection*{Client-Side Processing and Augmentation}

Given a batch of samples $X \in \mathbb{R}^{B \times C \times H \times W}$, where $B$ denotes the batch size, $C$ the number of channels, and $H$ and $W$ the image height and width, respectively, we first apply MixStyle-based augmentation to generate two additional augmented batches:
\begin{equation}
    X^{(1)} = M(X), \quad X^{(2)} = M(X),
\end{equation}

where $M(\cdot)$ represents the MixStyle module (described in Algorithm~2). This module interpolates channel-wise statistics (mean and standard deviation) between two randomly selected samples, effectively increasing diversity in the feature space and enhancing model robustness to domain shifts.

\subsubsection*{Representation Extraction and Prediction}

After MixStyle augmentation, the FedAlign framework extracts representations and generates predictions for each of the augmented batches. Let $Z$ represent the latent feature space. We decompose the model $f$ into two components:
\begin{equation}
    f = g \circ h,
\end{equation}
\begin{equation}
    Z = h(X), \quad Z^{(1)} = h(X^{(1)}), \quad Z^{(2)} = h(X^{(2)}),
\end{equation}
\begin{equation}
    \hat{Y} = g(Z), \quad \hat{Y}^{(1)} = g(Z^{(1)}), \quad \hat{Y}^{(2)} = g(Z^{(2)}).
\end{equation}

\subsubsection*{Loss Functions}

The final step involves computing the overall loss by integrating three key objectives that collectively ensure domain-invariant feature learning and robust prediction consistency:

\paragraph{Supervised Contrastive Loss ($L_{SC}$):}  
Encourages alignment of representations $(Z, Z^{(1)}, Z^{(2)})$ for samples sharing the same class label, thereby promoting discriminative yet domain-invariant features.

\paragraph{Representation Consistency Loss ($L_{RC}$):}  
Uses Mean Squared Error (MSE) to minimize the discrepancy between the original and augmented representations $Z$ and $\{Z^{(1)}, Z^{(2)}\}$, thus reinforcing representation stability under distribution shifts.

\paragraph{Jensen–Shannon Divergence ($L_{JS}$):}  
Enforces prediction consistency by minimizing the divergence between $\hat{Y}$ and $\{\hat{Y}^{(1)}, \hat{Y}^{(2)}\}$. This ensures that the model’s outputs remain reliable even after augmentation.

By integrating these alignment mechanisms, FedAlign drives the extraction of domain-invariant features and bolsters the global model’s capacity to generalize effectively across heterogeneous domains.

\subsection{Cross-Client Feature Augmentation with MixStyle}

\subsubsection*{MixStyle-Based Cross-Client Feature Augmentation}

To tackle the challenge of limited domain diversity in federated learning, we incorporate an enhanced version of MixStyle a computationally lightweight data augmentation strategy. By perturbing style information (e.g., color and texture), MixStyle effectively simulates additional, previously unseen domains, thus broadening the training data distribution and bolstering model robustness against domain shifts.

\subsubsection*{Style Mixing Mechanism}

\noindent Consider a batch of input samples $X = \{x_i \mid i = 1, \ldots, B\}$, where $B$ is the batch size. For each sample $x$, MixStyle computes the channel-wise mean $\mu(x)$ and standard deviation $\sigma(x)$ as:
\begin{align}
\mu(x)_c &= \frac{1}{HW} \sum_{h=1}^{H} \sum_{w=1}^{W} x_{c,h,w}, \\
\sigma(x)_c &= \sqrt{\frac{1}{HW} \sum_{h=1}^{H} \sum_{w=1}^{W} (x_{c,h,w} - \mu(x)_c)^2},
\end{align}
where $H$ and $W$ are the height and width of the feature map, and $c$ indexes the channels. Given two samples $x_i$ and $x_j$, the style statistics are interpolated as:
\begin{align}
\gamma_{\text{mix}} &= \lambda \cdot \mu(x_i) + (1 - \lambda) \cdot \mu(x_j), \\
\beta_{\text{mix}} &= \lambda \cdot \sigma(x_i) + (1 - \lambda) \cdot \sigma(x_j),
\end{align}
where $\lambda \sim \text{Beta}(\alpha, \alpha)$ is sampled from a Beta distribution. The augmented sample is then produced by:
\begin{align}
x_{\text{aug}} = \gamma_{\text{mix}} \cdot \frac{x_i - \mu(x_i)}{\sigma(x_i)} + \beta_{\text{mix}}.
\end{align}

\subsubsection*{Improvements in MixStyle}

We extend MixStyle with two key modifications that enhance its capacity to capture domain-invariant features:

\paragraph{Clustering} We group features into clusters according to their style statistics, thereby facilitating the learning of domain-invariant representations. By explicitly clustering features with similar style properties, the method gains a more nuanced view of diverse domain factors.

\paragraph{Probabilistic Sampling Weights} To further encourage diversity, we weight clusters based on feature variance. This adaptive sampling mechanism prioritizes challenging or underrepresented samples, improving the model’s robustness to domain shifts.

\subsubsection*{Diversity Enhancement}

By simulating styles from multiple domains, the enhanced MixStyle approach significantly diversifies the training data distribution. This is particularly valuable in heterogeneous federated learning settings, where local data often exhibit substantial variability. Ultimately, the broadened style space fortifies the global model against domain shifts, leading to more generalizable and reliable performance.

\subsection{Adversarial Training}
To further enhance domain-invariant feature learning, we incorporate adversarial training by employing a 
domain discriminator that distinguishes between original and augmented representations. 
Simultaneously, the feature extractor is optimized to minimize the discriminator’s ability to 
differentiate domains, thereby promoting domain invariance. The domain discriminator itself comprises 
fully connected layers with dropout, batch normalization, and non-linear activations, ensuring robust 
performance across feature dimensions. This adversarial mechanism effectively mitigates domain shift 
and bolsters generalization across diverse client data distributions.

\subsection{Representation Alignment}

To promote domain-invariant feature learning, FedAlign incorporates two complementary losses that align representations across original and augmented samples.

\subsubsection*{Supervised Contrastive Loss ($L_{SC}$)}

This component aligns features of samples sharing the same label, thereby improving the class-level coherence of the learned representations. Formally, for a batch index set $I = \{1, 2, \ldots, B\}$, we define:
\begin{equation}
    L_{SC} = \sum_{i \in I} -\frac{1}{|P(i)|} \sum_{p \in P(i)} \log \left( 
    \frac{\exp(\text{sim}(z_i, z_p) / \tau)}{\sum_{a \in A(i)} \exp(\text{sim}(z_i, z_a) / \tau)} 
    \right),
\end{equation}

where:
\begin{itemize}
    \item $P(i)$ is the set of indices for samples having the same label as $i$.
    \item $\text{sim}(z_i, z_p)$ indicates the cosine similarity between $z_i$ and $z_p$.
    \item $\tau$ is a temperature parameter used to control the concentration of the distribution.
\end{itemize}

By maximizing similarity for positive pairs $(z_i, z_p)$ while minimizing similarity for negative pairs, $L_{SC}$ encourages class-aligned and discriminative representations.

\subsubsection*{Representation Consistency Loss ($L_{RC}$)}

To further ensure stability and consistency in the feature space, we incorporate a Mean Squared Error (MSE) term between original and augmented representations:
\begin{equation}
    L_{RC} = \frac{1}{|\text{mix\_feat}|} \sum \| h(X) - h(X_{\text{aug}}) \|^2,
\end{equation}

where $h(\cdot)$ denotes the representation encoder, and $\text{mix\_feat}$ refers to the set of feature maps selected for MixStyle augmentations. By minimizing $L_{RC}$, the model maintains consistency in latent representations, even when subjected to domain-altering transformations.

\subsubsection*{Total Representation Alignment Loss}

We combine these objectives into a single representation alignment loss, which balances both discriminative class alignment and robust consistency:
\begin{equation}
    L_{RA} = L_{SC} + L_{RC}.
\end{equation}

Through this unified formulation, FedAlign learns domain-invariant and stable feature embeddings that enhance its ability to generalize effectively to unseen target domains.

\begin{figure*}[t]
    \centering
    \includegraphics[width=\textwidth, keepaspectratio]{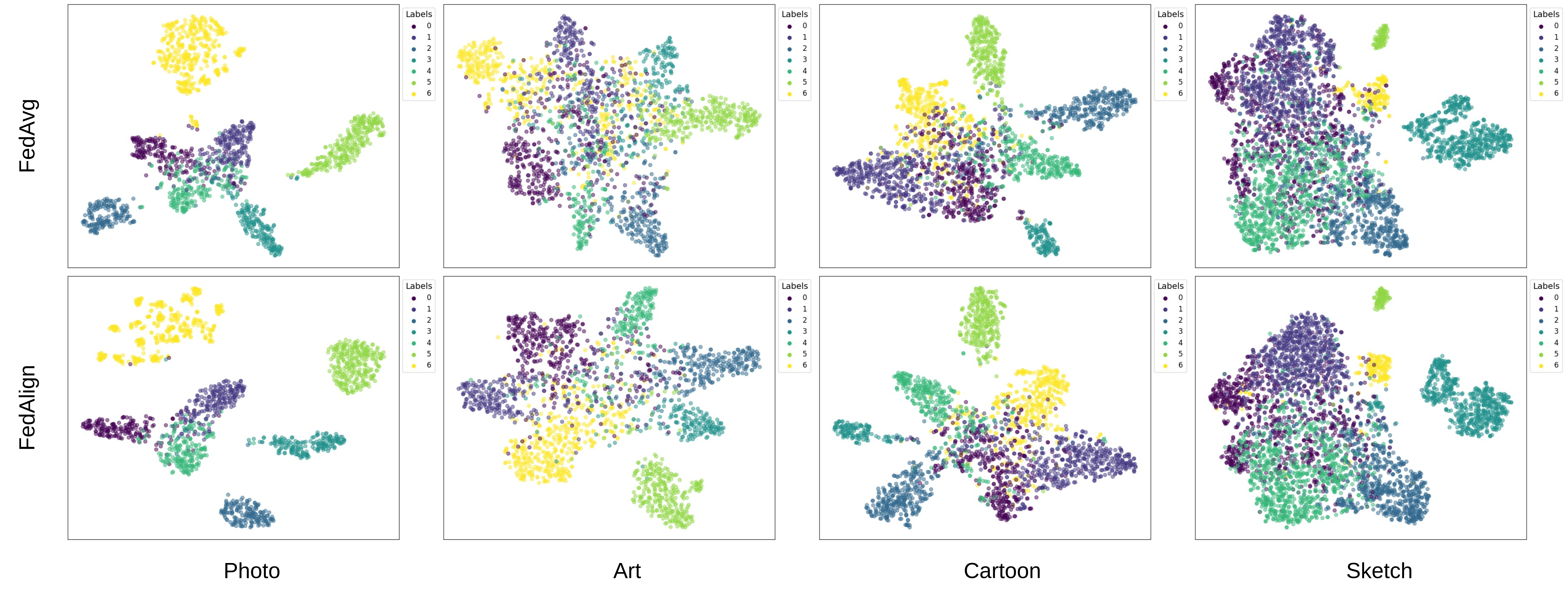}
    \caption{t-SNE visualization of the representation distribution using FedSR. The representations show domain-specific clusters with noticeable overlaps, highlighting the limitations of FedSR in learning robust domain-invariant features.}
    \label{fig:tsne}
\end{figure*}

% \begin{figure*}[t]
%     \centering
%     \includegraphics[width=\textwidth, keepaspectratio]{tsne_fedalign.png}
%     \caption{t-SNE visualization of the representation distribution using FedAlign. The representations form distinct and compact clusters, demonstrating improved domain generalization and alignment across domains compared to FedSR.}
%     \label{fig:tsnefedalign}
% \end{figure*}

\subsection{Prediction Alignment}

In addition to feature-level alignment, FedAlign imposes consistency on the model’s outputs through Jensen–Shannon (JS) Divergence, which measures the stability of predictions across original and augmented samples. Formally, for predictions $Y$, $Y^{(1)}$, and $Y^{(2)}$ (corresponding to $X$, $X^{(1)}$, and $X^{(2)}$, respectively), the JS Divergence loss is defined as:
\begin{equation}
    L_{JS} = \frac{1}{3} \left[ KL(Y \| \bar{Y}) + KL(Y^{(1)} \| \bar{Y}) + KL(Y^{(2)} \| \bar{Y}) \right],
\end{equation}

where:
\begin{itemize}
    \item $\bar{Y} = \frac{1}{3}(Y + Y^{(1)} + Y^{(2)})$ is the mean prediction distribution across the original and augmented samples.
    \item $KL$ denotes the Kullback–Leibler Divergence, quantifying how one probability distribution diverges from a second, reference distribution.
\end{itemize}

By enforcing prediction consistency, $L_{JS}$ encourages the network to produce stable outputs despite the domain-perturbing augmentations, thereby promoting robust and domain-invariant classification performance.

\subsection{Total Loss Function}

The final loss function for FedAlign combines the primary classification objective with both representation and prediction alignment terms:
\begin{equation}
    L = L_{CLS} + \lambda_1 (L_{SC} + L_{RC}) + \lambda_2 L_{JS},
\end{equation}

where:
\begin{itemize}
    \item $L_{CLS}$ is the cross-entropy loss for classification.
    \item $\lambda_1$ and $\lambda_2$ are hyperparameters that balance the influence of the representation and prediction alignment terms, respectively.
\end{itemize}

By integrating these complementary objectives, FedAlign fosters domain-invariant representations and stable predictions, culminating in a robust federated learning framework with strong generalization to unseen domains.

% \begin{table}[t]
%     \centering
%     \renewcommand{\arraystretch}{1.2} % Adjust row height
%     \setlength{\tabcolsep}{4pt} % Adjust column spacing
%     \adjustbox{max width=\columnwidth}{
%     \begin{tabular}{ccc|ccc}
%         \toprule
%         $\mathcal{L}_{RA}$ & $\mathcal{L}_{JS}$ & MixStyle & PACS & OfficeHome & miniDomainNet \\
%         \midrule
%         &  &  & 79.92 & 66.58 & 59.57 \\
%         \checkmark &  &  & 80.95 & 66.93 & 61.43 \\
%         & \checkmark &  & 80.52 & 67.34 & 60.82 \\
%         \checkmark & \checkmark &  & 81.11 & 67.58 & 61.83 \\
%         \midrule
%         &  & \checkmark & 80.32 & 67.01 & 60.12 \\
%         \checkmark &  & \checkmark & 81.23 & 67.49 & 61.57 \\
%         & \checkmark & \checkmark & 80.75 & 67.89 & 61.34 \\
%         \checkmark & \checkmark & \checkmark & \textbf{82.46} & \textbf{68.31} & \textbf{62.14} \\
%         \bottomrule
%     \end{tabular}
%     }
%     \caption{Ablation study results. The total number of clients is set to 6 and the upload ratio is set to 0.1.}
%     \label{tab:ablation_study}
% \end{table}

\begin{figure*}[t]
    \centering
    \includegraphics[width=\textwidth]{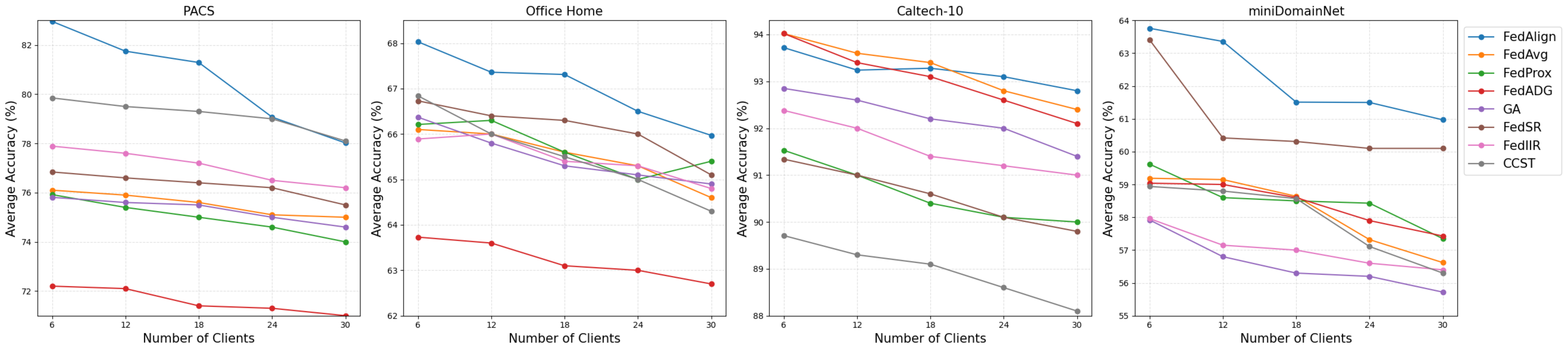}
    \caption{Average test accuracy (\%) versus the number of participating clients.}
    \label{fig:chartresults}
\end{figure*}

\section{Experiments}
\noindent \textbf{Datasets.}  
We evaluate \textbf{FedAlign} on four widely used domain generalization benchmarks, each offering distinct characteristics and posing unique challenges:

\begin{itemize}
    \item \textbf{PACS} \citep{li2017deeper}: This dataset contains 9,991 samples spread across four domains: Art Painting, Cartoon, Photo, and Sketch. Comprising 7 classes, PACS is known for its substantial inter-domain variability, making it a stringent testbed for domain generalization methods.
    \item \textbf{OfficeHome} \citep{venkateswara2017deep}: OfficeHome includes 15,588 samples from four domains: Art, Clipart, Product, and Real World, covering 65 categories. It is frequently employed in both domain adaptation and domain generalization tasks due to the diversity of object appearances arising from everyday office and home environments.
    \item \textbf{miniDomainNet} \citep{zhou2021domain}: A subset of DomainNet, miniDomainNet contains 140,006 images from four domains—Clipart, Infograph, Painting, and Real—and spans 126 categories. Its large-scale, heterogeneous nature presents significant challenges for learning domain-invariant representations.
    \item \textbf{Caltech (Caltech-101)} \citep{griffin2007caltech256}: Often referred to as Caltech-101, this dataset comprises 9,146 images across 101 object categories. Despite its relatively smaller size, its broad range of object classes allows for a robust evaluation of domain generalization strategies.
\end{itemize}

\noindent \textbf{Evaluation Protocol.}  
To thoroughly assess generalization performance, we employ the widely adopted leave-one-domain-out protocol. Specifically, for each dataset, one domain is designated as the test set while the remaining domains are collectively used as the training set. This procedure is repeated for every domain, ensuring that each serves once as the unseen target domain. By systematically testing across multiple distribution shifts, this protocol enables a comprehensive evaluation of the model’s ability to generalize to novel domains.

\noindent \textbf{Computational and Transmission Overhead.}  
Although sample statistics (e.g., mean and variance) are shared among clients in FedAlign, the corresponding computational and transmission overhead is minimal when compared to the cost of model training and communication. Furthermore, adversaries cannot reconstruct samples solely from these statistics, preserving data privacy. As demonstrated in Figure~5, FedAlign achieves superior performance with minimal upload ratios, underscoring its efficiency in both reducing communication overhead and mitigating privacy risks.

\noindent \textbf{Data Partitioning.}  
We follow the partitioning strategy presented in Section 3.1 of our overall methodology. Specifically, each dataset is split among a predefined number of clients, with variations in the composition of local training data across clients. This setup simulates realistic non-IID distributions commonly observed in federated learning environments, thereby providing a stringent assessment of each method’s robustness to data heterogeneity.

\noindent \textbf{Model Architecture.}  
All methods, including our proposed FedAlign, adopt MobileNetV3-Large as the backbone network. The final fully connected layer is employed as the classifier $g$, while the preceding layers collectively serve as the representation encoder $h$.

\noindent \textbf{Training Configuration.}  
The training proceeds for 10 communication rounds, with each client performing 3 local epochs during each round. We use the Adam optimizer, initializing the learning rate at 0.001 and decreasing it via cosine decay over the course of training for smoother convergence. For the supervised contrastive loss, we set the temperature parameter $\tau = 0.1$, balancing inter-class separability and intra-class coherence. Input images from PACS and OfficeHome datasets are resized to $224 \times 224$, whereas those from miniDomainNet are resized to $128 \times 128$.

\section{Experimental Results}

\subsection{Quantitative Performance and Comparative Analysis}

As shown in Table~\ref{tab:accuracy}, FedAlign consistently outperforms all baseline methods across the evaluated datasets, achieving the highest overall average accuracy. Notably, FedAlign also secures the top accuracy in each target domain for both the PACS and miniDomainNet benchmarks, underscoring its robust generalization capabilities.

Furthermore, we observe that most existing methods explicitly designed for \textit{Federated Domain Generalization (FDG)} often fail to maintain stable performance across different datasets; in some cases, they even lag behind classical federated learning approaches such as FedAvg and FedProx. This indicates that many current FDG algorithms may not sufficiently address the challenges of learning domain-invariant features under federated constraints, emphasizing the need for a more robust solution like FedAlign.

\subsection{Scalability and Robustness Analysis}

In addition to superior average accuracy, Figure~\ref{fig:chartresults} illustrates the resilience of FedAlign under varying client sizes, encompassing both small- and large-scale client settings. While the performance of all baseline methods deteriorates markedly as the number of participating clients increases, FedAlign maintains a consistent advantage. This robustness highlights FedAlign’s ability to adapt to diverse federated learning scenarios, effectively balancing scalability with state-of-the-art performance.

\subsection{Representation Distribution via t-SNE}

To further evaluate the effectiveness of the proposed FedAlign framework, we examine the distribution of learned representations using t-SNE visualizations. As shown in Figure~\ref{fig:tsne}, we compare the representation spaces across four domains—Photo, Art, Cartoon, and Sketch—under two settings: the baseline method (FedAvg, top row) and our FedAlign framework (bottom row). These visual comparisons provide valuable insights into the ability of FedAlign to learn domain-invariant features.

\subsubsection*{Distinct and Compact Clusters}

In challenging domains such as Photo and Art, FedAlign yields more distinct and compact clusters. This indicates that the framework effectively mitigates distributional gaps among diverse domains, suggesting stronger domain alignment than the baseline.

\subsubsection*{Improved Intra-Class Coherence}

Within each cluster, samples from the same class are more tightly grouped under FedAlign. This suggests a higher degree of feature alignment across clients and domains, translating to enhanced generalization performance.

\subsubsection*{Enhanced Inter-Class Separability}

FedAlign also achieves better separation between different classes, reducing overlap and confusion in the representation space. As a result, the learned features exhibit higher discriminative power, critical for robust domain generalization.

Overall, the compactness of clusters and the improved class separation observed in t-SNE plots confirm that FedAlign effectively handles domain shifts, thereby offering more robust and generalized feature representations compared to traditional federated learning baselines.

\subsubsection*{Implications for Domain Generalization}

The observed improvements in representation distribution underscore the efficacy of \textbf{FedAlign} in promoting feature alignment across diverse domains. By learning robust, domain-invariant features, FedAlign substantially boosts generalization performance, particularly when tackling previously unseen target domains. This enhanced resilience to domain shifts is crucial for real-world \textit{Federated Domain Generalization (FDG)} applications, where heterogeneity is often unavoidable. The t-SNE visualizations confirm that FedAlign successfully narrows the gaps between source domains while preserving strong predictive accuracy, thereby demonstrating its potential to handle challenging and heterogeneous federated environments.

\section{Conclusion}
In this paper, we present FedAlign, a novel framework for Federated Domain Generalization (FDG) that addresses the challenges of limited local data and client heterogeneity. It aims to significantly enhance model generalization by introducing an efficient cross-client feature extension module, that enriches and diversifies representations. Additionally, it employs a dual-stage alignment strategy targeting both feature representations and output predictions to robustly extract domain-invariant features. Extensive evaluations on multiple standard benchmark datasets demonstrate that our framework consistently outperforms state-of-the-art methods, delivering superior accuracy and strong scalability across varying client populations.

\bibliographystyle{named}
\bibliography{ijcai25}

\end{document}